\documentclass[twoside,11pt]{article}
\usepackage{automl2019}
\usepackage[utf8]{inputenc} %
\usepackage[T1]{fontenc}    %
\usepackage{hyperref}       %
\usepackage{url}            %
\usepackage{booktabs}       %
\usepackage{amsfonts}       %
\usepackage{nicefrac}       %
\usepackage{microtype}      %
\usepackage{graphicx}
\usepackage{subcaption}
\usepackage{amsmath}

\jmlrheading{Zal\'an Borsos, Andrey Khorlin and Andrea Gesmundo}

\ShortHeadings{Transfer NAS: Knowledge Transfer between Search Spaces with Transformer Agents}{Z. Borsos, A. Khorlin and A. Gesmundo}
\firstpageno{1}

\begin{document}

\title{Transfer NAS: Knowledge Transfer between Search Spaces with Transformer Agents}

\author{\name Zal\'an Borsos\thanks{This work was done during an internship at Google AI.} \email zalan.borsos@inf.ethz.ch \\
       \addr Department of Computer Science, ETH Zurich
       \AND
       \name Andrey Khorlin \email akhorlin@google.com \\
       \addr Google Research, Brain Team
       \AND
       \name Andrea Gesmundo \email agesmundo@google.com \\
       \addr Google Research, Brain Team}

\maketitle

\begin{abstract}
Recent advances in Neural Architecture Search (NAS) have produced state-of-the-art architectures on several tasks. NAS shifts the efforts of human experts from developing novel architectures directly to designing architecture search spaces and methods to explore them efficiently. The search space definition captures prior knowledge about the properties of the architectures and it is crucial for the complexity and the performance of the search algorithm. However, different search space definitions require restarting the learning process from scratch. We propose a novel agent based on the Transformer that supports \emph{joint training} and \emph{efficient transfer} of prior knowledge between multiple search spaces and tasks.

\end{abstract}

\section{Introduction}
    
    Neural Architecture Search falls under the umbrella of AutoML and targets the automated design of neural networks. Successful approaches include methods based on reinforcement learning (RL) \citep{zoph2016neural, baker2016designing}, evolutionary algorithms \citep{miller1989designing, real2018regularized, elsken2018efficient}, Bayesian optimization \citep{bergstra2013making, golovin2017google} and gradient-based methods \citep{liu2018darts}. 
    Random search \citep{bergstra2012random} has also proven to be a competitive approach to NAS, especially when the search space is constrained \citep{li2019random}. 
    
    At the heart of these successes lie well-designed \emph{search space definitions}. When confronted with a new  search space, the majority of the methods restart their search procedure from scratch. This is in stark contrast to how a human practitioner would reuse prior knowledge obtained on similar search spaces. Motivated by this intuition, in this work, we propose a novel agent that efficiently transfers prior knowledge to previously unseen search spaces and tasks, leading faster discovery of good sets of hyperparameters. Our agent's architecture is based on the \emph{Transformer} \citep{vaswani2017attention}, and is designed to support joint training over multiple search spaces and tasks. When transferring to NAS-Bench-101 \citep{ying2019bench}, our agent provides a speedup of factor 3-4 over the competing methods. 

\section{Related Work}

    Following \cite{elsken2018neural}, NAS methods can be characterized by 1) the search space definition, 2) the search strategy and 3) the performance estimation strategy they employ. The search space definition is an abstract representation of all the conceivable architectures and it captures prior knowledge about the properties of the architectures at different granularities (e.g., cell-based design \citep{zoph2018learning}). A common design approach is the fixed-length, linear search space, where the set of hyperparameters to tune is static. Complex dependencies between choices can be modeled by \emph{conditional search spaces} \citep{swersky2014raiders, elsken2018neural, de2018neural}, that support variable length sequences of decisions.
    
    The search space is explored by a search strategy, which faces the exploration-exploitation dilemma: it should find good architectures early, but it should also avoid converging to suboptimal choices.
    In this work, we focus on the RL-based approach to NAS \citep{zoph2016neural, baker2016designing}, where the search space is explored by an agent whose actions define the architectural parameters. The performance estimation strategy consists of training the child model with the chosen parameters on the task at hand. The measure of the child model's quality on the validation set is presented as the reward signal to the agent. The goal of the agent is to find good sets of hyperparameters that maximize the expected reward. RL-based NAS has been successful at designing RNN cells \citep{zoph2016neural}, convolutional blocks \citep{zoph2018learning} and optimizers \citep{bello2017neural}.
    
    Transfer learning for hyperparameter search has a long history that mainly focuses on transfer between tasks using the same search space. Transfer methods relying on Bayesian optimization \citep{bardenet2013collaborative, swersky2013multi, golovin2017google, perrone2018scalable} have shown promising results, but they need to address several specific issues, such as handling large dimensional spaces, the dependence of runtime complexity on the number of trials, and importantly, the change in scale of the objective function on different tasks. In the context of RL-based NAS, \citet{wong2018transfer} proposed an agent that efficiently transfers prior knowledge to unseen tasks relying on task embeddings and parameter reuse. Whereas training on conditional search spaces has been explored in several works  \citep{swersky2014raiders, feurer2015initializing, de2018neural}, to the best of our knowledge, transfer between search spaces has received no attention.
    
\section{Methods}

    In this work, we focus on the RL-based approach to NAS, and propose a novel self-attention-based agent  relying on the Transformer that supports joint training over multiple search spaces and tasks and can efficiently transfer knowledge to new search spaces and tasks. We opt for a policy gradient-based approach where the transfer is achieved through employing \emph{shared embeddings} for the states, tasks and actions as well as \emph{shared model parameters} between different search spaces and tasks. We now discuss our design choices in detail.
    
    \paragraph{Why policy gradients?} Several methods for hyperparameter search rely on a quality function $f: \mathbb{D} \times \mathbb{H} \mapsto \mathbb{R} $. This function maps a dataset $D \in \mathbb{D}$ and the set of hyperparameters $h \in \mathbb{H}$ to the quality score $f(D, h)$, which is the quality metric of the generated model trained on $D$ with parameters $h$. However, $f$ can be non-smooth in $D$, i.e., the quality metric can scale very differently for different datasets even for the same set of hyperparameters. This  poses challenges when considering transfer to new tasks and search spaces  \citep{bardenet2013collaborative}, since the scale difference must be captured. Similar issues arise in RL-based approaches to NAS using Q-learning or value function estimation.
    On the other hand, policy gradient methods circumvent the usage of such quality function by directly optimizing the  policy for sampling hyperparameters. Thus, they have significant advantage in transfer, since the optimal policy transfers to similar setups without requiring scale adaptation. This motivates policy-based RL as our default choice for transfer in NAS.

    \paragraph{Joint training and transfer.} In order to learn and transfer between multiple search spaces $SS=\{ss_1,...,ss_n\}$  and tasks $T=\{t_1,...,t_m\}$ efficiently, a viable option is to perform joint training over the available search space and task combinations $A=\{ (ss_i, t_j) \mid ss_i \textup{ valid on } t_j \}$. However, performing this operation is not immediate due to the changing search space definition over optimization rounds. 
    
\begin{figure}[h]
        \centering
        \includegraphics[width=0.9\textwidth]{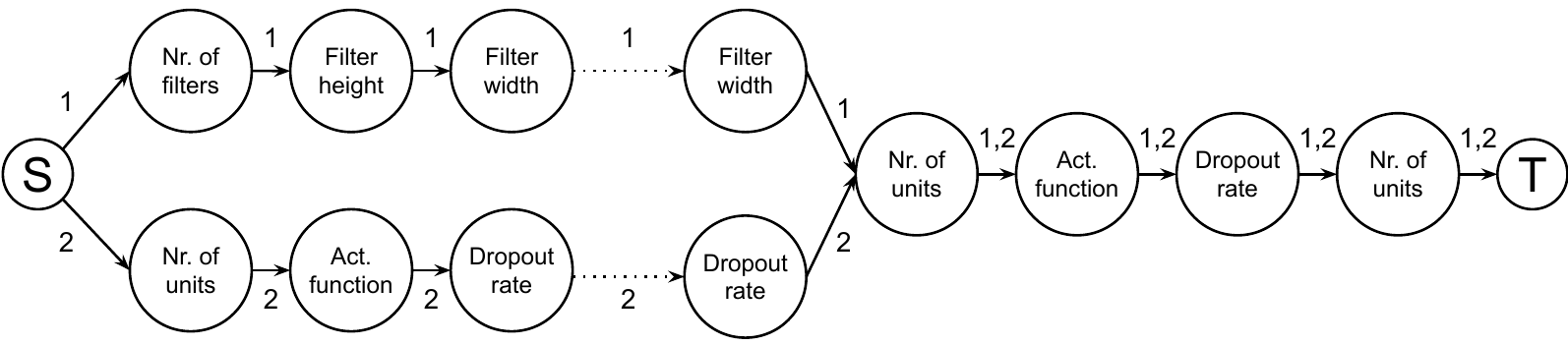}
        \vspace{-2mm}
        \caption{Merged conditional search spaces for a CNN and a FFN.}
        \label{fig:merged-search-space}
        \vspace{-5mm}
    \end{figure}

    A first idea would be to merge the search spaces into a fixed-length, linear space. However, the length of the sequence can be prohibitively large for large $|SS|$. To remedy this, we propose to \emph{merge} different search spaces into a larger conditional search space, where the available actions are not only conditioned on past actions, but also on the sampled search space and task pair $(ss_i, t_j) \sim \textup{Uniform}(A)$. This idea is illustrated in Figure \ref{fig:merged-search-space}, where a merged conditional search space of hyperparameters of a convolutional neural net ($ss_1$) and a feedforward net ($ss_2$) is shown. $S$ and $T$ denote the start and terminal state, whereas the annotations on the edges show their availability conditioned on the chosen search space. Notice how the conditional search space allows for merging the states for the fully connected layer of the CNN and the last layer of the FFN. Now, the joint training can be efficiently performed in this conditional search space.
    
    \begin{figure}
        \centering
        \begin{subfigure}[b]{0.4\linewidth}
            \centering
            \includegraphics[width=\linewidth]{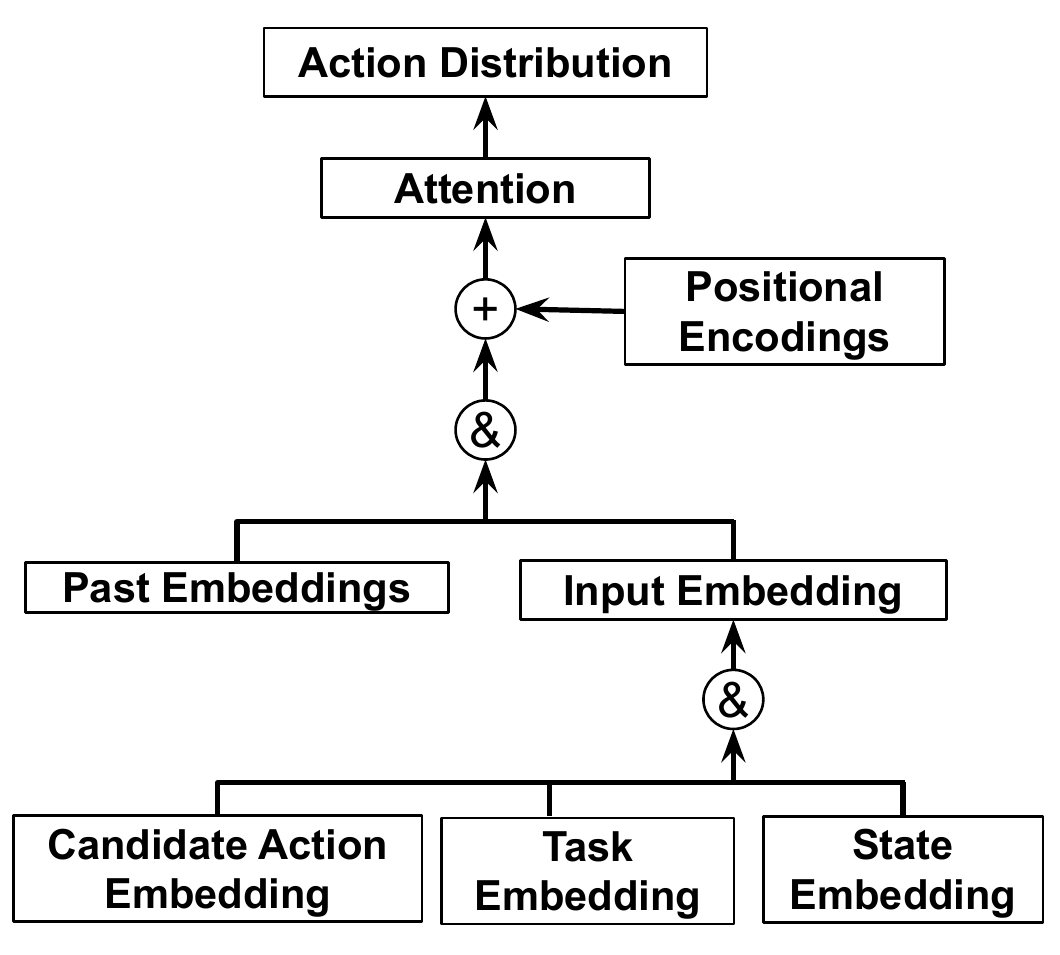}
            \caption{The overall architecture.}
            \label{fig:architecture}
        \end{subfigure}
        \hspace{7mm}
        \begin{subfigure}[b]{0.4\linewidth}
            \centering
            \includegraphics[width=\linewidth]{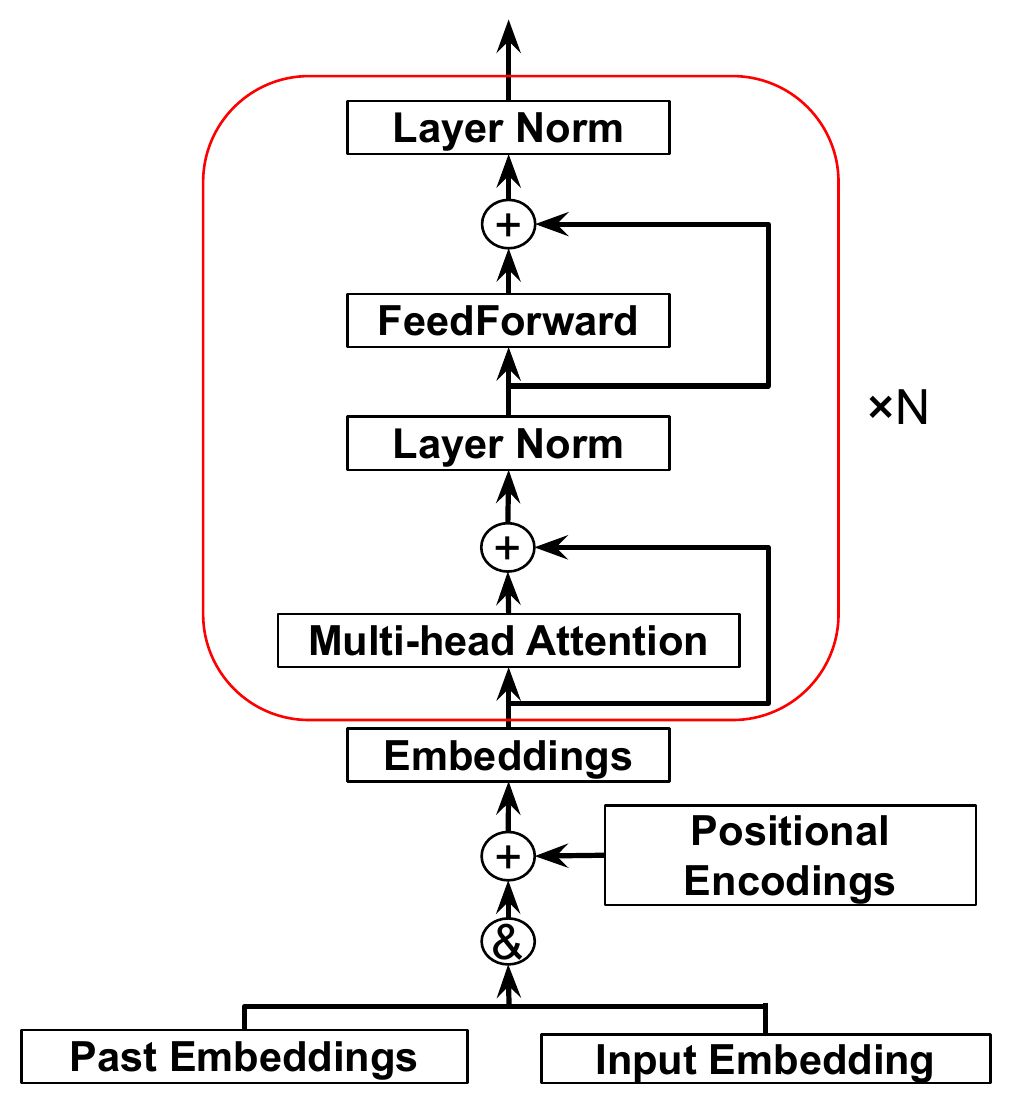}
            \caption{The attention architecture.}
            \label{fig:attention-layer}
        \end{subfigure}%
        \caption{The proposed architecture for the agent based on the Transformer.}
        \vspace{-5mm}
    \end{figure}
    
    \paragraph{Attention-based agent.} Models based solely on attention have recently shown promising results in translation \citep{vaswani2017attention}, question answering and language inference \citep{devlin2018bert}. So far unexplored in the context of RL-based NAS, a self-attention-based architecture for the agent possesses several \emph{desirable properties} for transfer and joint training on multiple search spaces and tasks.
    
    In the approach of \cite{zoph2016neural}, the agent relies on an LSTM-based RNN for representing the policy. The RNN is conditioned on the action history through its hidden state, as well as on the embedding of the last action received as input.
    However, as illustrated in Figure \ref{fig:merged-search-space}, the merged conditional search space can share several states between different search spaces, which renders \emph{conditioning on last action}  challenging, e.g., at states that are reachable in multiple search spaces. The conditioning on the last action is further complicated by the fact that  different search spaces might define the sequence of states in a different order.
    Moreover, these states can have \emph{long-term dependencies} between them, requiring the RNN to be extended with attention to past anchor points \citep{zoph2016neural}. The range of the long-term dependencies may vary significantly from one search space to another. 
    
    All of the issues mentioned above are remedied once we consider a purely \emph{self-attention-based architecture} for the agent. This way, the agent can attend to past actions up to \emph{arbitrarily long} distances while also allowing to possibly retain the relative position of actions through learned positional encodings. Thus, inspired by the Transformer, we propose our agent architecture as shown in Figure \ref{fig:architecture}, where $+$ denotes addition, $\&$ denotes concatenation and $N$ represents the number of hidden layers in the attention. The attention architecture in Figure \ref{fig:attention-layer} is identical to the encoder of the Transformer.

    We employ embeddings for the states, tasks and actions that are \emph{shared} in the merged conditional space as follows. The state embeddings are reused based on the state name (e.g., "Nr. of filters" in Figure \ref{fig:merged-search-space}). We focus on categorical actions, where each state has different embedding for each action. While the action embeddings are unique per state, they are reused through states with identical names. The conditioning on the search space is achieved through the action embeddings in state $S$, which act  as search space embeddings. 
    
    At each step, the input embedding is a concatenation of the candidate action, task and state embeddings. The past embeddings consist of the concatenation of the input embeddings corresponding to previously sampled actions. We thus obtain as many input embeddings as available actions at a given step, each of which is concatenated with the past embeddings, and passed through the attention. For each action, we retain the last embedding from the output of the attention layer and we map it to logits via a state-specific fully connected layer. The logits are then mapped to parameters of the categorical sampling distribution via the softmax function. The parameters of the attention are shared through all search spaces and tasks.

    \paragraph{Training and transfer considerations.} We employ learned positional encodings, single-head and two-layer attention.  All the embedding sizes are fixed to 8. The training is performed with REINFORCE  together with entropy regularization and Priority Queue Training \citep{abolafia2018neural} using Adam \citep{DBLP:journals/corr/KingmaB14} with learning rate $5\cdot10^{-4}$. More details are provided in the Appendix. The parameters were chosen based on the double XOR toy task presented in Section \ref{subsec:double-xor} and reused for other experiments. 
    
    When transferring to new search spaces and tasks with possibly unseen states and actions, the agent's parameters and the relevant embeddings are reloaded, the new embeddings are randomly initialized and the training is resumed. However, we must ensure that the unseen actions are explored. We find that entropy regularization alone is not a viable option in the first rounds of transfer as strong regularization might cancel the benefits of reusing prior knowledge. As a remedy, at time $t$, we mix the distribution defined by policy $\pi$ over available actions $A_t=\{a_t^1, ...,a_t^k\}$ with the uniform distribution using the mixing coefficient $\gamma$,
    \begin{equation}
    \pi'(a_t^i| a_1,...a_{t-1}) = (1-\gamma) \cdot \pi(a_t^i| a_1,...a_{t-1}) + \gamma / k, \qquad \textup{ for all } i\in \{ 1,...,k\},
    \end{equation}
    where $\gamma$ is decayed linearly from $0.1$ to 0 over to course of the first 100 sampling steps. 
    Similar smoothing approaches are applied in algorithms such as the EXP3 \citep{auer2002nonstochastic}.

\section{Experiments}
    \subsection{Double XOR} \label{subsec:double-xor}
    In order to showcase the behavior of our Transformer-based agent and its benefits over an LSTM-based agent similar to the one presented in \cite{zoph2016neural}, we consider the toy task consisting of a chain  with 7 states $\{ S_1, ... ,S_7\}$ and binary actions $\{ a_1, ... ,a_7\}\in \{0,1\}^7$, where the reward is defined as $a_1 \oplus a_5 \wedge a_3 \oplus a_7 $. This interleaved XOR task is designed to imitate long-term dependencies between states and to render conditioning on last sampled actions challenging.
    \begin{figure}[h]
        \centering
        \vspace{-3mm}
        \begin{minipage}[b]{0.48\linewidth}
            \centering
            \includegraphics[width=\linewidth]{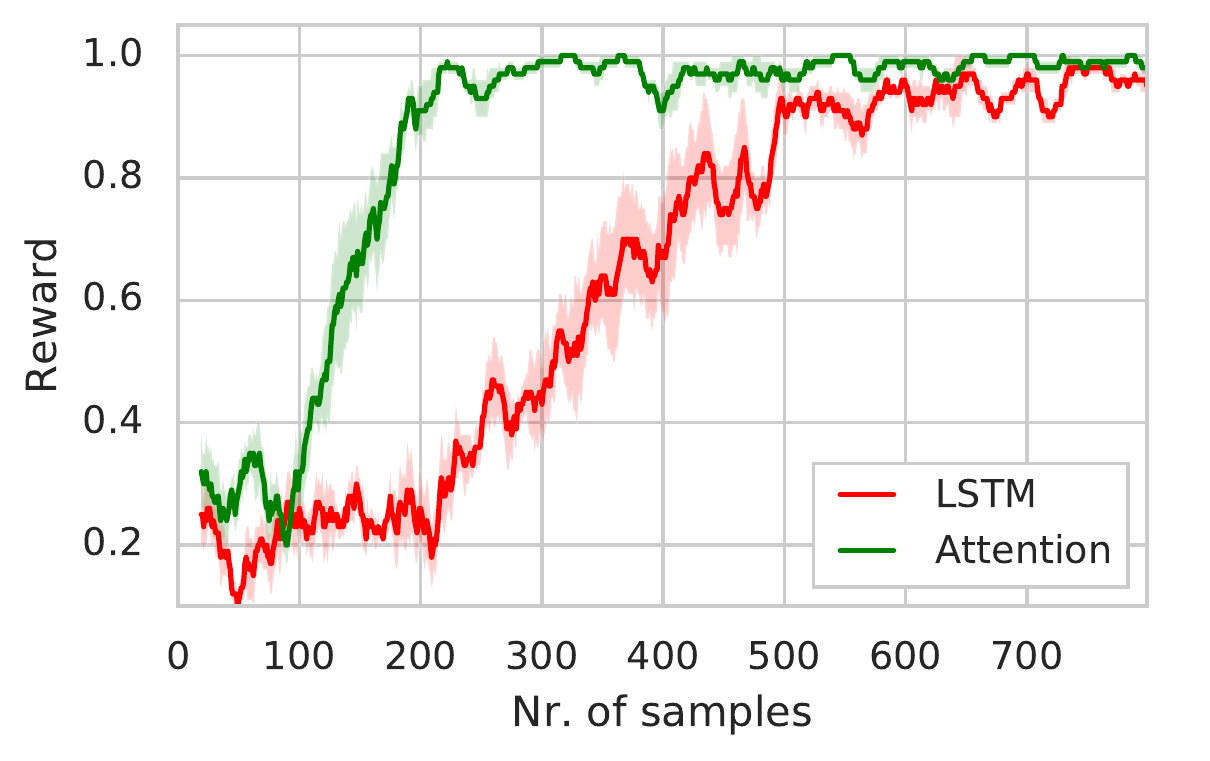}
            \vspace{-6mm}
          \caption{The moving average of the rewards for the double XOR.}
            \label{fig:double-xor-rewards}
        \end{minipage}
        \hfill
        \begin{minipage}[b]{0.48\linewidth}
            \centering
            \includegraphics[width=\linewidth]{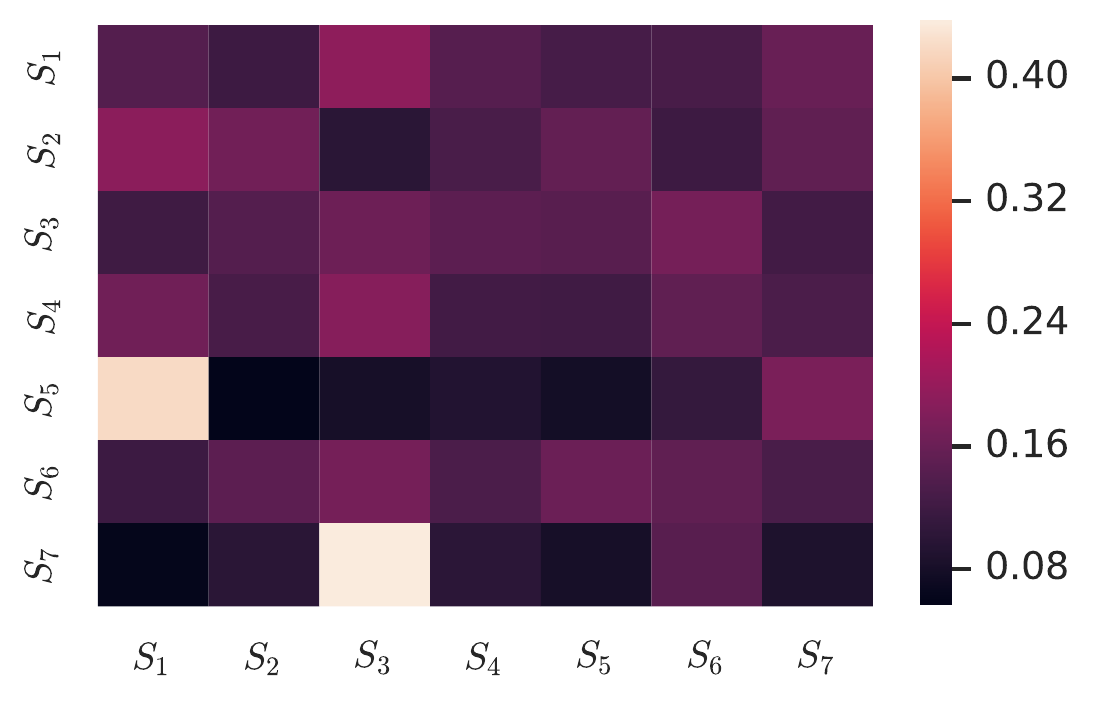}
            \vspace{-6mm}
            \caption{The attention weights of the first attention layer obtained for the double XOR.}
            \label{fig:attention-weights}
        \end{minipage}
        \vspace{-3mm}
    \end{figure}
    
    We train the attention-based and the LSTM-based agent on this task, with comparable number of parameters, where the hyperparameters where tuned separately for the methods with the goal of earliest convergence. We plot the moving average of the rewards of 5 experiment replicas in Figure \ref{fig:double-xor-rewards}, where the shaded areas represent $68\%$ confidence intervals. The results shows the clear advantage of the attention-based agent in capturing the long-term dependencies. In Figure \ref{fig:attention-weights}, we plot the attention map of the first hidden layer for verification. This map mirrors the definition of the rewards, i.e., the action in $S_5$ and $S_7$ strongly depend on the past actions taken in $S_1$ and $S_3$, respectively.  
    
    \subsection{Transfer between Search Spaces on NAS-Bench-101}
    NAS-Bench-101 \citep{ying2019bench} is a public architecture search benchmark, that allows to test NAS agents on the task of designing a CNN cell. Each cell is defined by 7 vertices %
    and at most 9 edges. One of the three available operations is selected for each vertex: 1) 3x3 conv-BN-ReLU, 2) 1x1 conv-BN-ReLU, 3) 3x3 max-pool. NAS-Bench provides information about results of training all possible architectures, serving as fast surrogate for benchmarking NAS algorithms without the cost of training and evaluating computationally expensive CNNs. 
    
    We measure the benefits of transferring the agent across search spaces by first training on two subsets of the full NAS-Bench search space and then transferring to the full search space. Following  \cite{ying2019bench}, we define our full search space as a chain with 26 states, where the first 21 states have binary actions that indicate the presence of each possible edge. The last 5 states choose the operations for the vertices. The two subspaces that we jointly  train on are defined only over 6 vertices (vertex 6 is removed) and for each vertex a subset of operations is available: operations $\{1, 2\}$ and $\{1, 3\}$, respectively.
    
    We compare the transferred agent to: 1) same agent trained from scratch, 2) random search and 3) Regularized Evolution (RE) \citep{real2018regularized}. Refer to the Appendix for details about the hyperparameters setup.  All the compared agents are configured to keep sampling until a valid configuration is generated.  We also measure the regret, defined as the difference between the test accuracy of the best possible architecture in the search space and the test accuracy of the best models. 
    
\begin{figure}[h]
        \centering
        \includegraphics[width=0.98\textwidth]{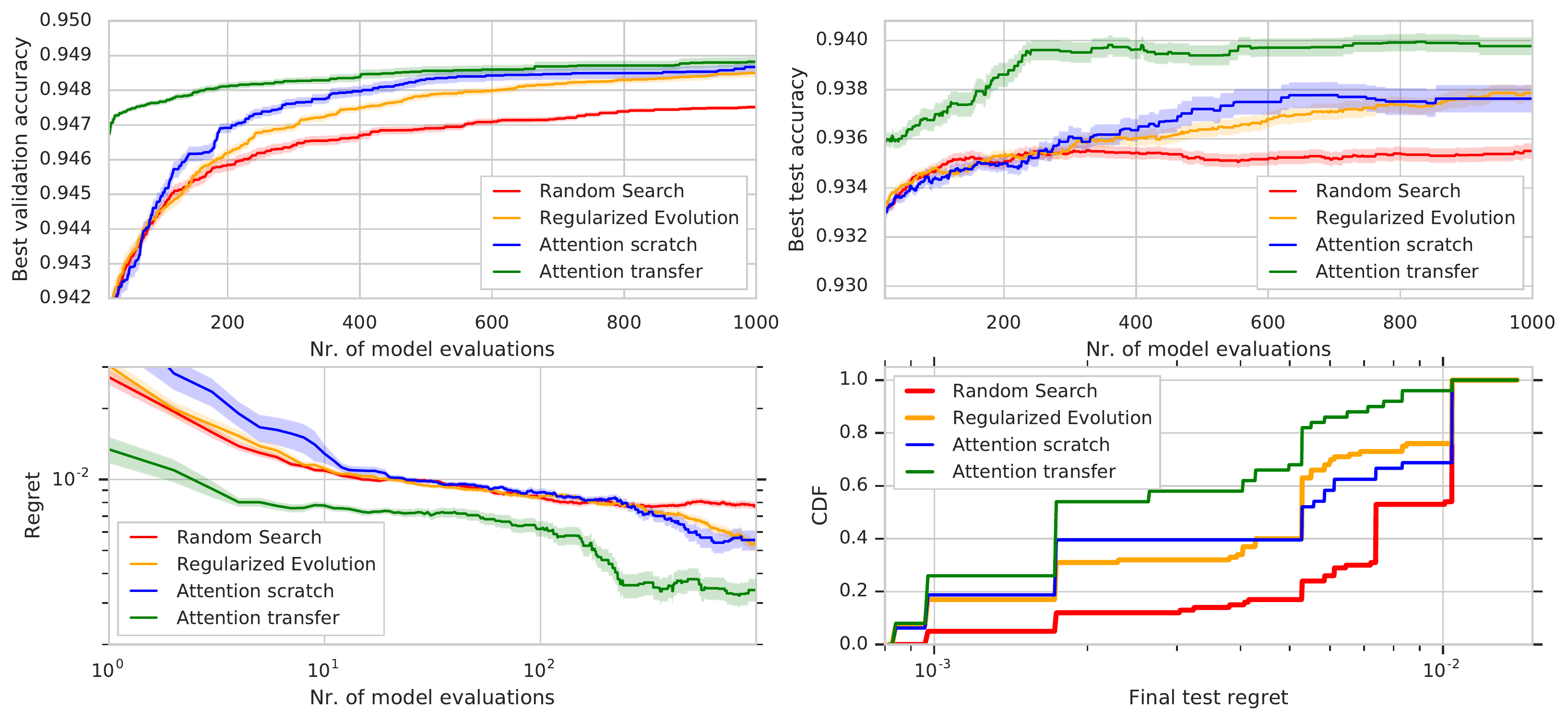}
        \vspace{-2mm}
        \caption{Transfer learning on NAS-Bench-101.}
        \label{fig:nasbench-transfer}
        \vspace{-3mm}
    \end{figure}

    Figure \ref{fig:nasbench-transfer} shows the results averaged over 50 experiment replicas. Transfer reduces the number of trials needed to reach a certain validation accuracy by a factor of 3-4 compared to training from scratch and to RE. Similar trends are reflected also by the test accuracy curves of the chosen models. When trained from scratch, our agent performs on par with RE.
    \vspace{-2mm}
    \paragraph{Conclusion} We proposed a novel agent for RL-based NAS that relies on self-attention-based architecture and supports efficient knowledge transfer between multiple search spaces and tasks. When transferring to NAS-Bench-101, our agent provides 3-4 factor speedup over the competing methods. 
    \newpage

\bibliography{bibliography} 

\newpage
\appendix

\section{Training Method}

The training is performed with REINFORCE  using entropy regularization and Priority Queue Training (PQT)  \citep{abolafia2018neural}. We use undiscounted rewards which are only observed once the full sequence of actions is generated.

For REINFORCE, we employ baselines for each search spaces and task pair that are calculated using exponential moving average over the rewards. For PQT, we consider the top $K=25$ experiences for each search space and task pair. For a uniformly drawn pair ($ss_i$, $t_j$), we sample a sequence of actions $(a_1, ..., a_M) \sim \pi_\theta$, where $\theta$ denotes all model parameters including the embeddings, and estimate the gradient of the loss from three components:

\begin{equation}
\begin{gathered}
\nabla_\theta J(\theta) = (R(a_1, ..., a_m) - b_{ss_i, t_j}) \cdot \sum_{m=1}^M \nabla_\theta \log \pi_\theta(a_m | a_1,...,a_{m-1}; ss_i, t_j) \\
\nabla_\theta L(\theta) =  \sum_{m=1}^{M'} \nabla_\theta \log \pi_\theta(\tilde{a}_m | \tilde{a}_1,...,\tilde{a}_{m-1}; ss_i, t_j) \\
\nabla_\theta H(\theta) = - \frac{1}{M}\sum_{m=1}^M \sum_{a \in A_m} \nabla_\theta \left( \pi_\theta(a | a_1,...,a_{m-1}; ss_i, t_j) \log \pi_\theta(a | a_1,...,a_{m-1}; ss_i, t_j) \right)
\end{gathered}
\end{equation}
where $(\tilde{a}_1, ..., \tilde{a}_{M'})$ is a uniformly sampled sequence over the top $K$ experiences for seen for ($ss_i$, $t_j$) and $A_m$ is the set of valid actions at position $m$. The final gradient is formed as
\begin{equation}
\nabla_\theta L(\theta) + \lambda_{\textup{PQT}}\nabla_\theta L(\theta) + \lambda_{\textup{ENT}} \nabla_\theta H(\theta),
\end{equation}
where $\lambda_{\textup{PQT}} = 5$ and $\lambda_{\textup{ENT}} = 0.15$ where chosen based on the double XOR toy task. 

\section{NAS-Bench-101 Experimental Setup}
For selecting the hyperparameters of our agent, we tuned them on the double XOR toy task presented in Section \ref{subsec:double-xor} and reused these parameters for the NAS-Bench-101 both for training from scratch and transfer.

For Regularized Evolution, following \cite{ying2019bench}, we define two types of mutation: edge mutation and operator mutation. In each mutation round, we randomly choose between these two types of mutations. If an edge mutation should be performed, we randomly sample one out of the 21 possible edges and flip it. For operator mutations, we sample one out of 5 operator positions and then sample a new operator for it. The population size is fixed to 100 and the sample size was chosen from $\{2, 2^2, 2^3, 2^4, 2^5, 2^6\}$ based on final test validation of the best model found after performing the evolution, and it was set to $2^3$.    
\end{document}